\documentclass[journal]{IEEEtran}

\usepackage{times}
\usepackage{epsfig}
\usepackage{graphicx}
\usepackage{amsmath}
\usepackage{amssymb}
\usepackage{dsfont}
\usepackage{amsthm}
\usepackage{algorithm}
\usepackage{algorithmic}
\usepackage{bm}

\usepackage[numbers,sort]{natbib}

\newtheorem{remk}{Remark}

\newtheorem{defn}{Definition}
\newtheorem{lema}{Lemma}

\ifCLASSINFOpdf
\else
\fi

\begin{document}
%
\title{On Better Exploring and Exploiting Task Relationships in Multi-Task Learning: Joint Model and Feature Learning}
%
%
%

\author{Ya~Li,
        Xinmei~Tian,~\IEEEmembership{Member,~IEEE,}
        Tongliang Liu,
        and~Dacheng~Tao,~\IEEEmembership{~Fellow,~IEEE}
\thanks{Ya Li and Xinmei Tian are with the department of Electronic Engineering and Information Science, University of Science and Technology of China. }
\thanks{Tongliang Liu and Dacheng Tao are with the University of Technology, Sydney.}
\thanks{\textcopyright 2018 IEEE. Personal use of this material is permitted.
Permission from IEEE must be obtained for all other uses, in any current or future media, including reprinting/republishing this material for advertising or promotional purposes, creating new collective works, for resale or redistribution to servers or lists, or reuse of any copyrighted component of this work in other works.}
}

%
%

\markboth{IEEE TRANSACTIONS ON NEURAL NETWORKS AND LEARNING SYSTEMS}%
{Shell \MakeLowercase{\textit{et al.}}: Bare Demo of IEEEtran.cls for Journals}
%



\maketitle

\begin{abstract}
Multi-task learning (MTL) aims to learn multiple tasks simultaneously through the interdependence between different tasks. The way to measure the relatedness between tasks is always a popular issue. There are mainly two ways to measure relatedness between tasks: common parameters sharing and common features sharing across different tasks.  However, these two types of relatedness are mainly learned independently, leading to a loss of information. In this paper, we propose a new strategy to measure the relatedness that jointly learns shared parameters and shared feature representations. The objective of our proposed method is to transform the features from different tasks into a common feature space in which the tasks are closely related and the shared parameters can be better optimized. We give a detailed introduction to our proposed multi-task learning method. Additionally, an alternating algorithm is introduced to optimize the non-convex objection. A theoretical bound is given to demonstrate that the relatedness between tasks can be better measured by our proposed multi-task learning algorithm.  We conduct various experiments to verify the superiority of the proposed joint model and feature multi-task learning method.
\end{abstract}


%
\IEEEpeerreviewmaketitle

\section{Introduction}
%
%
%
%
\IEEEPARstart{S}{ingle-task} learning learns different tasks separately by ignoring the intrinsic relatedness between different tasks. However, multi-task learning can well prevent this drawback by jointly measuring the interdependence between different tasks. The performance of all tasks is supposed to be improved with additional information provided by the relationship between tasks.  Consider the merits of multi-task learning, it has been applied to various research areas, for example, web image search \cite{wang1}, video tracking \cite{Xue39}, disease prediction \cite{zhang2}, and relative attributes learning \cite{chen3}.  \par

MTL makes the assumption that tasks have some intrinsic relatedness. Consequently, proper measurement of task relatedness will benefit the learning of tasks and improve the performance of each other.  Conversely, improper relatedness measurement introduces noise and degrades the performance. Recently, researchers have given substantial attention to measuring task relatedness. Existing algorithms mainly use two methods to measure the relatedness between tasks: shared common models/parameters \cite{Evgeniou7,Xue8,Kai9,rai10, liu32, lee33} and shared common feature representations \cite{argyriou4, jebara5, Lapin6, meier34, obozinski35, hou44, hou45}. MTL of sharing common models/parameters (multi-task model learning) makes the assumption that models of different tasks have something in common in their parameters.  MTL of sharing common feature representations (multi-task feature learning) assumes that related tasks share a subset of features to measure relatedness. \par

 Both multi-task model learning and multi-task feature learning suffer from their own defects. They only consider one aspect of task relatedness. For example, the relatedness is directly captured in the original feature space by multi-task model learning. However, considering the noise and complexity of features in real-world datasets, task relatedness measured by the original features may not be obvious. As a result, the performance of multi-task model learning may degrade. Multi-task feature learning prevents this drawback by learning a common subset of feature representations. However, it ignores the relatedness between model parameters. We develop a new multi-task model and feature joint learning method in this paper that can successfully explore task relatedness. Our model learns a common feature space shared by different tasks in which the relatedness between tasks is maximized. Consequently, the common models can be better measured jointly.

The objective function is formulated as a non-convex problem and an alternating algorithm is proposed to optimize it. Additionally, we present sound theoretical analysis to prove the better ability of measuring task relatedness with our joint model and feature learning method. Various experimental results are reported to  demonstrate the effectiveness of our proposed method ,especially on tasks with shared features or shared models. \par

The remainder of this paper is organised as follows. In Section \ref{sec:relatedworks}, we briefly review previous multi-task learning works. In Section \ref{sec:cmt}, we give a detailed derivation and optimization algorithm of our proposed method. Section \ref{sec:theroAnaly} derives a theoretical error bound to demostrate the merits of our proposed algorithm. Experimental results are reported in Section \ref{sec:exp} with conclusions and future work given in Section \ref{sec:conclusion}.

\section{Related work} \label{sec:relatedworks}

In recent years, researchers have paid much attention to multi-task learning.  Compared to single-task learning, its effectiveness has been demonstrated through theoretical analysis in many works \cite{baxter17, Ben24, Maurer25, Ando26, Cong43}. For example, a novel inductive bias learning method was proposed by Baxter \cite{baxter17}. This work derived explicit bounds, demonstrating that learning multiple related tasks within an environment potentially achieves substantially better generalization than does learning a single-task. Ben-David and Schuller proposed a useful concept of task relatedness \cite{Ben24} to derive a better generalization of error bounds.  Maurer et al. applied the dictionary learning and sparse coding to multi-task learning and introduced a generalization bound by measuring the hypothesis complexity \cite{Maurer25}. Ando and Zhang made assumption that all tasks shared a common structure and showed a reliable estimation of shared parameters between tasks when the number of tasks was large. \cite{Ando26}.\par

With more extensive applications of multi-task learning, some single-task learning algorithms have been extended to multi-task learning framework. For example, some works extended Bayesian method into multi-task learning methods with the assumption that the models of tasks are indeed related \cite{Heskes27, Heskes28}. Hierarchical Bayesian models can be learned by sharing parameters as hyperparameters at a high level. The relatedness between tasks can also be measured by deep neural networks, such as sharing nodes or layers of the network \cite{Caruana29}. As one of the most popular single-task learning methods, SVM has been studied in many multi-task learning works \cite{Evgeniou7, Jun30, Evgeniou31, jebara5, Cai40, Cong41, Pareto42}. Jebara proposed a multi-task learning method using maximum entropy discrimination based on the large-margin SVM \cite{jebara5}. Zhu et al. propose an infinite latent SVM for multi-task learning \cite{Jun30}. It combines the large-margin idea with a nonparametric Bayesian model to discover the latent features for multi-task learning. \par

The most difficult aspect of multi-task learning is simultaneously measuring the relatedness between tasks and keeping the individual characteristics. Multi-task model learning and multi-task feature learning are two main categories of multi-task learning methods. For multi-task model learning, Xue et al. proposed two different forms of MTL problem using a Dirichlet process based statistical model and developed efficient algorithms to solve the proposed methods \cite{Xue8}. Evgeniou and Pontil introduced a multi-task learning model by minimizing a regularized objection similar to support vector machines \cite{Evgeniou7}. This work assumed that all tasks shared a mean hyperplane with a particular offset on their own. A nonparametric Bayesian model was proposed by Rai and Daume \cite{rai10} to capture task relatedness under the assumption that parameters shared a latent subspace. The dimensionality of the subspace is automatically inferred by the proposed model. For  the category of multi-task feature learning, Argyriou et al. developed a convex MTL method for learning shared features between tasks \cite{argyriou4}. The learned common features were regularized by a L21-norm to control the dimensionality of the latent feature space. Jebara proposed a general multi-task learning framework using large-margin classifiers. Three scenarios are discussed:  multit-task feature learning, multi-task kernel combination and graphical multi-task model \cite{jebara5}. To improve the efficiency of multi-task learning on high-dimensional problems, a novel multi-task learning method was proposed by learning low-dimensional features of tasks jointly  \cite{Lapin6}. \par

Recently, the defects of measuring task relatedness in traditional multi-task learning methods have been widely discussed. The assumptions that all tasks are related through sharing common parameters or common features are usually not suitable for real-world multi-task learning problems. Considering the defects of such assumptions, a number of works \cite{kang11, Gong13, Chen14, Jalali12} have been proposed to improve the performance of multi-task learning. For example, Kang et al. learned a shared feature representation across tasks while simultaneously clustering the tasks into different groups \cite{kang11}.  Chen et al. proposed a robust multi-task learning method that learned multiple tasks jointly while simultaneously finding outlier tasks \cite{Chen14}. Another robust multi-task feature learning method was proposed by Gong et al. \cite{Gong13}. This model was similar to the method in \cite{Chen14}. This work decomposed the weight matrix into two components and imposed the group Lasso penalty on both components. The group Lasso penalty was imposed on the row of the first component for capturing the shared features between relevant tasks, and the same group Lasso penalty was imposed on the column of the second component to find outlier tasks. Another work \cite{Jalali12} proposed a dirty model for multi-task learning by utilizing a idea similar to \cite{Gong13,Chen14}. The model uses both block-sparse regularization and element-wise sparsity regularization to capture the true features used for each task. Block-sparse regularization learned the shared features across tasks, and element-wise regularization guaranteed that some features were used for some tasks but not all. These works can be divided into two categories: task clustering and outlier task finding.  \par
However, these works only consider one aspect of task relatedness: either shared features or shared parameters. In this paper, we consider the shared features and shared parameters simultaneously to overcome the problems in existing multi-task learning methods. The relatedness can be better modeled in our multi-task learning framework, especially when both feature relatedness and model relatedness exist between tasks.

\begin{figure*}[!tb]
\label{fig CMT Illustration}
\begin{center}
\includegraphics[width=0.8\textwidth]{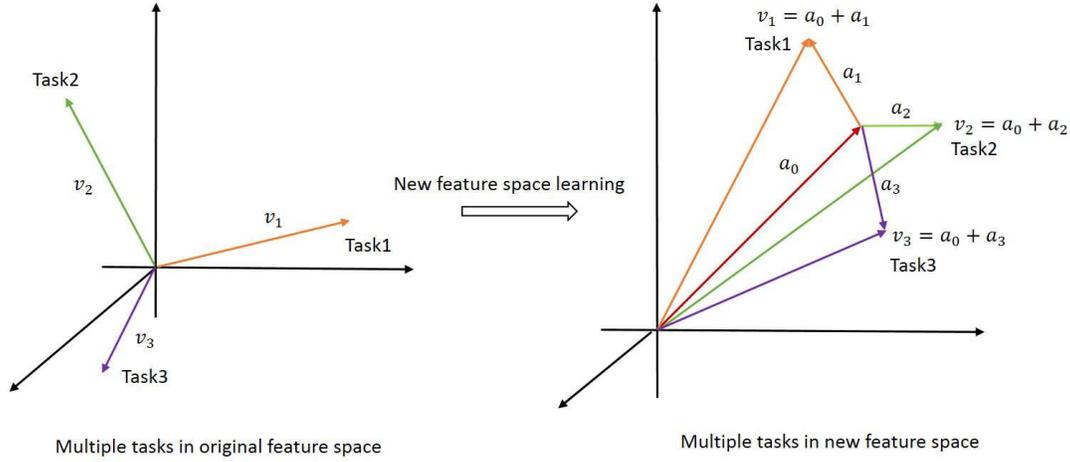}
\end{center}
\caption{Framework of our proposed multi-task learning method. In the original feature space, tasks have weak relatedness. We aim to map the data into a new space, therefore all tasks can be more closely related and share a common hyperplane in this new feature space.}
\end{figure*}

\section{Multi-task model and feature joint learning} \label{sec:cmt}
We introduce our newly proposed multi-task learning method specifically in this section. We first show the objective optimization problem and then convert the non-convex problem into a convex formulation. An efficient optimization algorithm is given at the end of this section. \par
The idea of our proposed method is illustrated in Figure 1. There are three related tasks in the original feature space. However, the interdependence between them is not as strong as assumed in multi-task learning due to the noise and complexity of feature representation. It may lead to bad performances of multi-task learning in the original feature space. In our work, we transform the original feature space into a new feature space, in which different tasks are tightly related and are possible to share a common hyperplane $a_0$. The specific characteristic of task $t$ is represented by an offset $a_t$.

\subsection{Non-convex objective}
Suppose we have $T$ different tasks and each task $t$ is related with a dataset $D_t$ which can be formulated as follows:
$$D_t = \{(x_{t1},y_{t1}), (x_{t2},y_{t2}), ..., (x_{tm_t},y_{tm_t})\},$$
where $m_t$ is the number of data samples in task $t$.  $x_{ti} \in \mathds(R)^d$ and $y_{ti} \in \mathds{R}$ are the corresponding feature representation and output of sample $i$ in task $t$.
In this work, we consider to learn $T$ different linear functions to predict the output given the input feature representation in each task:
\begin{equation}
f_t(x_{ti})=v_t^Tx_{ti}\approx y_{ti},
\end{equation}
where $t \in \{1, 2,...,T\}$.  Single-task learning methods treat these $T$ linear functions as separate tasks and just utilize the data information from each task. Consequently, it ignores the interdependence between tasks which may provide more valuable information about the distribution of training data. Consider the drawbacks of single-task learning algorithm, multi-task learning are proposed to uncover the relatedness between tasks and gain performance improvement of all tasks. The improvement is expected to be obvious especially with a small amount of training data. The relatedness between tasks can provide more additional information in such situation.   \par
In this work, we first learn a feature mapping matrix $U \in \mathds{R}^{d \times d}$ to get better relatedness between tasks in the new feature space:
\begin{equation}
\label{eq_4}
f_t (x_{ti})=\langle a_t+a_0,U^T x_{ti}\rangle,
\end{equation}
note $v_t = a_t + a_0$. $a_0$ is the shared central hyperplane in the new feature space and $a_t$ represents the offset of task $t$ to maintain its own characteristic. The learned feature mapping matrix $U$ is supposed to guarantee the assumption that all tasks share a central hyperplane with an offset in the new feature space. With the above formulation, our objective function of multi-task learning can be formulated as:

\begin{equation}
\label{proposed}
\min \limits_{V,a_0,U} \sum\limits_{t=1}^{T}\sum\limits_{i=1}^{m_t} \emph{l}\left(y_{ti}, \langle v_t,U^Tx_{ti}\rangle\right)+ \frac{\gamma}{T} \|V - a_0*\mathbf{1} \|_{2,1}^{2} + \beta\|a_0\|_2^2,
\end{equation}
where $V=[v_1, v_2, ..., v_T]$ and $\mathbf{1}$ represents a vector of all ones. Noting that $v_t = a_t + a_0$, we can reformulate problem (\ref{proposed}) with $A=[a_1, a_2, ..., a_T]$ as :
\begin{equation}
\label{changed}
\min \limits_{A,a_0,U} \sum\limits_{t=1}^{T}\sum\limits_{i=1}^{m_t} \emph{l}\left(y_{ti}, \langle a_t + a_0, U^Tx_{ti}\rangle\right)+ \frac{\gamma}{T} \|A\|_{2,1}^{2} + \beta\|a_0\|_2^2 .
\end{equation}
The third regularization term in problem (4) denotes the square of the L2-norm of vector $a_0$ which aims to measure the smoothness and complexity of the central hyperplane. The second regularization term is square of the  L21-norm of matrix $A$ which can be explicitly expressed as $\| A \|_{2,1} = (\sum_{i=1}^d \| a^i \|_2)$. $a^i$ denotes the $i$-th row of matrix $A$. The L21-norm guarantees that all tasks share a subset of common features and the sparsity of shared features. The first term is the loss function which measures the error between ground truth and predicted results. \par

There are three main differences between our proposed formulation and the formulation proposed in \cite{argyriou4}. First, the learning ability of feature mapping matrix $U$ has some limitation due to its orthogonal property. However, such limitations are ignored by the method proposed in \cite{argyriou4}. It is more reasonable to share a subset of common features around $a_0$ instead of a fix point at the origin. The proposed method can well prevent the limitation of orthogonal matrix $U$ by selecting features around a more robust point $a_0$. Second, the proposed method considers the task relatedness of both features and model parameters. However, the method proposed in \cite{argyriou4} just uncovers the shared common features across tasks leading to loss of information between related models. These tasks are treated independently when learning their model parameters in the learned new feature space. Third, it is more challenging to solve an optimization problem learning both of common features and common model parameters.  \par

The proposed objective function is  non-convex.  To briefly show the non-convexity of the problem, we give a counter example. Assuming that all the variables are scalars, it is easy to show that proposed objective is non-convex. It is usually difficult to get an optimal solution of a non-convex objective. Instead, we convert the non-convex objective into an equivalent convex problem. And an alternating algorithm is proposed to solve it in the following sections.

\subsection{Conversion to an equivalent convex optimization problem}
\newtheorem{theorem}{Theorem}

For simple optimization, the non-convex optimization problem (\ref{changed}) is converted to an equivalent convex problem in this section.

\begin{theorem}
\label{theorem 1}
The non-convex problem (\ref{changed}) can be equivalently converted to a convex optimization problem as follows:

\begin{eqnarray}\label{convex}
&\min \limits_{W,w_0,D} & \sum\limits_{t=1}^{T}\sum\limits_{i=1}^{m_t}  l\left(y_{ti}, \langle w_t+w_0,x_{ti}\rangle\right) \\
&&+ \frac{\gamma}{T} \sum\limits_{t=1}^T\langle w_t,D^+w_t\rangle + \beta\langle w_0,w_0\rangle, \nonumber \\
&\text{s.t.} & trace(D) \leq 1, range(W)\subseteq range(D),  D \in S_+^d. \nonumber
\end{eqnarray}

Suppose  $(\hat{W},\hat{w_0},\hat{D})$ is an optimal solution of convex problem (\ref{convex}), the corresponding optimal solution $(\hat{A}, \hat{a_0}, \hat{U})$ of non-convex problem (\ref{changed}) can be formulated as $\hat{A}=\hat{U}^T\hat{W}, \hat{a_0}=\hat{U}^T\hat{w_0}$ and the columns of $\hat{U}$ form an orthonormal basis of eigenvectors of $\hat{D}$. Additionally, suppose ($\hat{A}$, $\hat{a_0}$, $\hat{U}$) forms an optimal solution of non-convex problem (\ref{changed}), the corresponding optimal solution of convex problem (\ref{convex}) can be formulated as $\hat{W} = \hat{U}\hat{A}, \hat{w_0}=\hat{U}\hat{a_0}$, and $\hat{D} = \hat{U}Diag(\frac{\|\hat{a}^i\|_2}{\|\hat{A}\|_{2,1}})^d_{i=1}\hat{U}^T$.
\end{theorem}

Note that $trace(D) = \sum_{i=1}^d D_{ii}$  and $D \in S_+^d$ indicates that $D$ is a positive semidefinite symmetric matrix. $range(W)$ represents a set of vectors $\{x \in \mathds{R}^n: x = Wz,$ for some $ z \in \mathds{R}^T \}$. $D^+$ denotes the pseudoinverse of matrix $D$. $Diag(a_0)_{i=1}^d$ is a diagonal matrix and the vector $a_0$ forms the diagonal elements.

To show the convexity of problem (\ref{convex}), an additional function is introduced as $f:\mathds{R}^d\times S^d\rightarrow \mathds{R}\bigcup \{ +\infty\}$ which can be explicitly formulated as:

\begin{equation}
f(w,D) = \left\{
\begin{aligned}
& w^TD^+w \quad  if \quad D\in S_+^d \quad and \quad w \in range(D) \\
& +\infty
\end{aligned}
\right.
\end{equation}

With the additional function, problem (\ref{convex}) is equal to minimizing the sum of $T$ additional functions plus the loss term and the term $\langle w_0,w_0 \rangle$ in problem (\ref{convex}), subjected to the trace constraints. Its rightness can be guaranteed by the equality between the $T$ constraints $w_t\in range(D)$ and the constraint $range(W)\in range(D)$. The loss term in problem  (\ref{convex}) is the sum of loss function $l$, which is convex for $(w_t, w_0)$ and a linear map, therefore it is convex. Additionally, the term $\langle w_0, w_0 \rangle$ and the trace constraint is also convex. To show the convexity of problem (\ref{convex}), it is sufficient to show that $f$ is convex. The details of $f$ being convex can be found in  \cite{argyriou4}.

\subsection{An optimization algorithm}
An alternating optimization algorithm is proposed to optimize problem (\ref{convex}) corresponding to parameters $(W,w_0 )$ and $D$ in this section. Additionally, the final optimal solution of problem (\ref{changed}) can be obtained according to Theorem \ref{theorem 1}.

We first optimize problem (\ref{convex}) with respect to parameters $(W, w_0)$ by fixing matrix $D$. The optimization problem can be separated into $T$ different tasks with a fix $D$ in  \cite{argyriou4}. Comparing with the optimization of the objective in \cite{argyriou4}, the optimization of our newly proposed objective function is more challenging because of the shared parameter $w_0$. It cannot be viewed as $T$ independent optimization problems. Our objective can be formulated as:

\begin{eqnarray}\label{step1}
&\min \limits_{W,w_0} & \sum\limits_{t=1}^{T}\sum\limits_{i=1}^{m_t}  \emph{l}\left(y_{ti}, \langle w_t+w_0,x_{ti}\rangle\right) \\
 &&+ \frac{\gamma}{T} \sum\limits_{t=1}^T\langle w_t,D^+w_t\rangle + \beta\langle w_0,w_0\rangle, \nonumber \\
 &\text{s.t.} & trace(D) \leq 1, range(W)\subseteq range(D), D \in S_+^d. \nonumber
\end{eqnarray}

The loss function used in our work is a least square loss which is the same as that used in previous works. To solve problem (\ref{step1}), we introduce some additional variables. Note that $X_t =[x_{t1},x_{t2},...,x_{tm_t}] \in \mathds{R}^{d \times m_t}$ which denotes a data matrix of task $t$ and the corresponding output of task $t$ is represented as $Y_t=[y_{t1},y_{t2},...,y_{tm_t}]^T\in \mathds{R}^{m_t}$. $M$ denotes the sum of amount of data points from all $T$ tasks:
$$M = m_1 + m_2 +...+m_T.$$

Let $X = bdiag(X_1, X_2, ...,X_T)\in \mathds{R}^{dT \times M}$ and $Y = [Y_1^T, Y_2^T, ..., Y_T^T]^T \in \mathds{R}^{M}$. $bdiag\{X_1, X_2,...,X_T\}$ denotes a block diagonal matrix and its diagonal entries are data from the $T$ tasks. $Y$ denotes the outputs of all data belonging to the $T$ different tasks. Note $D_0 = bdiag(\underbrace{D,D,...,D}_T) \in \mathds{R}^{dT \times M}$, $W_0 = [\underbrace{w_0^T, w_0^T,...,w_0^T}_T]^T \in \mathds{R}^{dT}$ and $W_1 = [w_1^T, w_2^T,...,w_T^T]^T \in \mathds{R}^{dT}$. \par

Problem (\ref{step1}) can be reformulated as
\begin{equation}
\label{eq_all}
\min \limits_{W_1, W_0} \|Y-X^T(W_1+W_0)\|_2^2 + \frac{\gamma}{T} W_1^TD_0^+W_1+ \beta w_0^Tw_0.
\end{equation}

Note that  $W_0 = I_0 \times w_0$ with  $I_0 = [\underbrace{I,I,...,I}_T]^T \in \mathds{R}^{dT \times d}$ and $I \in \mathds{R}^{d \times d}$ denotes an identity matrix. We can reformulate problem (\ref{eq_all}) as a L2-norm regularized regression problem with some additional variables. Note that $Z_1 = \sqrt{\frac{\gamma}{T}} (D_0^+)^{\frac{1}{2}}W_1$, and let $Z_2 = \sqrt{\beta}w_0$. Then, $W_1 = \sqrt{\frac{T}{\gamma}}(D_0^+)^{-\frac{1}{2}}Z_1$ and $W_0 = \sqrt{\frac{1}{\beta}}I_0Z_2$. $(D_0^+)^{\frac{1}{2}} = bdiag(\underbrace{(D^+)^{\frac{1}{2}},(D^+)^{\frac{1}{2}},...,(D^+)^{\frac{1}{2}}}_T )$ and $(D_0^+)^{-\frac{1}{2}} = bdiag(\underbrace{(D^+)^{-\frac{1}{2}},(D^+)^{-\frac{1}{2}},...,(D^+)^{-\frac{1}{2}}}_T )$. We have

\begin{equation}
\begin{aligned}
& \frac{\gamma}{T} W_1^TD_0^+W_1+ \beta w_0^Tw_0  = [Z_1^T, Z_2^T][Z_1^T, Z_2^T]^T=Z^TZ \\
& W_1 + W_0  = [\sqrt{\frac{T}{\gamma}}{(D_0^+)}^{-\frac{1}{2}}, \sqrt{\frac{1}{\beta}}I_0][Z_1^T, Z_2^T]^T = MZ,
\end{aligned}
\end{equation}
note that $M= [\sqrt{\frac{T}{\gamma}}{(D_0^+)}^{-\frac{1}{2}}, \sqrt{\frac{1}{\beta}}I_0]$ and $Z = [Z_1^T, Z_2^T]$. Consequently, the above problem is reformulated as the following standard L2-norm regularized problem:

\begin{equation}
\min_Z \|Y-X^TMZ\|_2^2+Z^TZ.
\end{equation}
The solution can be explicitly expressed as the following:
\begin{equation}
Z = (M^TXX^TM+I)^{-1}M^TXY.
\end{equation}
Additionally, we need to optimize problem (\ref{convex}) with respect to  matrix $D$ by fixing parameters  $(W,w_0)$. The objective can be simply formulated as the following:

\begin{eqnarray}\label{step2}
& \min \limits_{D} &\sum \limits_{t=1}^{T}\langle w_t, D^+w_t\rangle,  \\
& \text{s.t.} &D \in S_+^d, trace(D) \leq 1, range(W) \subseteq range(D). \nonumber
\end{eqnarray}
The optimal solution is explicitly shown as (the details can be found in \cite{argyriou4}):
\begin{equation}
\hat{D} = \frac{(WW^T)^{\frac{1}{2}}}{trace(WW^T)^{\frac{1}{2}}}.
\end{equation}

\section{Theoretical Analysis} \label{sec:theroAnaly}
For better understanding the merits of our method, a generalization bound of the non-convex problem (\ref{changed}) is analysed in this section. We first reformulate the problem by converting the two soft constraints  $\frac{\gamma}{T}\|A\|_{2,1}^2$ and $\beta\|a_0\|_2^2$ into hard ones as the following:

\begin{eqnarray}\label{soft}
&\min \limits_{a_t,a_0,U,\varepsilon_1,\varepsilon_2}&\sum_{t=1}^{T}\sum_{i=1}^{m_t}l\left(y_{ti},\left<a_t+a_0,U^Tx_{ti}\right>\right)+\varepsilon_1+\varepsilon_2,\nonumber\\
&\text{s.t.}&\gamma\frac{1}{T}\|A\|_{2,1}^2\leq\varepsilon_1,\\
&&\beta\|a_0\|_2^2\leq\varepsilon_2.\nonumber
\end{eqnarray}

The demonstration of the equality between problem (\ref{soft}) and problem (\ref{changed}) can be found in  \cite{vainsencher2011sample}, and $\varepsilon_1$, $\varepsilon_2$ are of order $\mathcal{O}(1)$ . Denote that  $\varepsilon_1=\varepsilon_2=\mathcal{O}(1)$, the above problem can be formulated as follows:

\begin{eqnarray}\label{soft1}
&\min \limits_{a_t,a_0,U}&\sum_{t=1}^{T}\sum_{i=1}^{m_t}l\left(y_{ti},\left<a_t+a_0,U^Tx_{ti}\right>\right),\nonumber\\
&\text{s.t.}&\|A\|_{2,1}^2\leq\mathcal{O}\left(\frac{T}{\gamma}\right),\\
&&\|a_0\|_2^2\leq\mathcal{O}\left(\frac{1}{\beta}\right).\nonumber
\end{eqnarray}

Consequently, we analyse the problem with hard constraints instead. We derive a generalization bound of the proposed problem following a similar way to that of \cite{mehta2013sparsity} by setting $\varepsilon=1$:
\begin{eqnarray}\label{soft2}
&\min \limits_{a_t,a_0,U}&\sum_{t=1}^{T}\sum_{i=1}^{m_t}l\left(y_{ti},\left<a_t+a_0,U^Tx_{ti}\right>\right),\nonumber\\
&\text{s.t.}&\|A\|_{2,1}^2\leq\frac{T}{\gamma},\\
&&\|a_0\|_2^2\leq\frac{1}{\beta}.\nonumber
\end{eqnarray}

The loss function $l$ is supposed to satisfy the following Lipschitz-like condition, to simplify the analysis of the upper bound of the generalization error. This has been also used in \cite{mohri2012foundations}.

\begin{defn}
A loss function $l$ is $c$-admissible with respect to the hypothesis class $H$ if there exists a $c\in\mathbb{R}_+$, where $\mathbb{R}_+$ denotes the set of non-negative real numbers, such that for any two hypotheses $h, h'\in H$ and example $(x,y)\in\mathcal{X}\times\mathbb{R}$, the following inequality holds:
\[|l(y,h(x))-l(y,h'(x))|\leq c|h(x)-h'(x)|.\]
\end{defn}
We can have:
\begin{theorem}\label{main}
Suppose $B$ is the upper bound of loss function $l$, such that $l(y,f(x))\leq B$. And the loss function $l$ is $c$-admissible corresponding to the linear function class. For any optimal solution $(A, a_0, U)$ of problem (\ref{changed}), by replacing the hard constraints $\|A\|_{2,1}^2\leq\frac{T}{\gamma}$ and $\|a_0\|_2^2\leq\frac{1}{\beta}$ with soft constraints $\gamma\frac{1}{T}\|A\|_{2,1}^2$ and $\beta\|a_0\|_2^2$, and for any $\delta>0$, we have the following results with probability of at least $1-\delta$:

\begin{eqnarray*}
&&E_x\sum_{t=1}^{T}\sum_{i=1}^{m_t}l\left(y_{ti},\left<a_t+a_0,U^Tx_{ti}\right>\right) \\
&&-\sum_{t=1}^{T}\sum_{i=1}^{m_t}l\left(y_{ti},\left<a_t+a_0,U^Tx_{ti}\right>\right)\leq\\
&& 2c\left(\sqrt{\frac{T}{\gamma}}+\sqrt{\frac{1}{\beta}}\right)\sqrt{\sum_{t=1}^{T}m_tS(X_t)}+3B\sqrt{\frac{\sum_{t=1}^{T}m_t\ln(\frac{2}{\delta})}{2}},
\end{eqnarray*}
where $S(X_t)=\text{tr}\left(\hat{\Sigma}(x_t)\right)=\frac{1}{m_t}\sum_{i=1}^{m_t}\|x_{ti}\|_2^2$ is the empirical covariance for the observations of the $t$-th task. Letting $m_1=\ldots=m_T=m$ and $\|x_t\|_2\leq r,t=1,\ldots,T$, with a probability of at least $1-\delta$, we have
\begin{eqnarray*}
&&\frac{1}{T}\sum_{t=1}^{T}E_xl\left(y_{t},\left<a_t+a_0,U^Tx_{t}\right>\right)\\
&&-\frac{1}{T}\sum_{t=1}^{T}\frac{1}{m}\sum_{i=1}^{m}l\left(y_{ti},\left<a_t+a_0,U^Tx_{ti}\right>\right)\\
&&\leq \frac{2cr}{\sqrt{\gamma m}}+\frac{2cr}{\sqrt{\beta mT}}+3B\sqrt{\frac{\ln(2/\delta)}{2mT}}.
\end{eqnarray*}
\end{theorem}

\begin{remk}
The first term $\frac{2cr}{\sqrt{\gamma m}}$ in Theorem \ref{main} is the generalization bound related to the learning of matrix $A$ and the second term $\frac{2cr}{\sqrt{\beta mT}}$ corresponding to $a_0$. This theoretical result demonstrates that the learning of shared hyperplane $a_0$ is of order  $\mathcal{O}(\sqrt{1/mT})$ and it can be better learned with more tasks. Consequently, our proposed multi-task joint learning method can perform better than single-task learning methods. Additionally, $a_0$ is encouraged to be larger with the constraints of $\|A\|_{2,1}$ and the utility of feature mapping matrix $U$. Thus, the generalization bound of our proposed method have a faster convergence than the method proposed in \cite{argyriou4}, which demonstrates the efficiency of our method.
\end{remk}

The proof of Theorem \ref{main} is given in Appendix A.

\begin{figure}[!tb]
	\label{fig_weightA0}
	\begin{center}
		\includegraphics[width=0.4\textwidth]{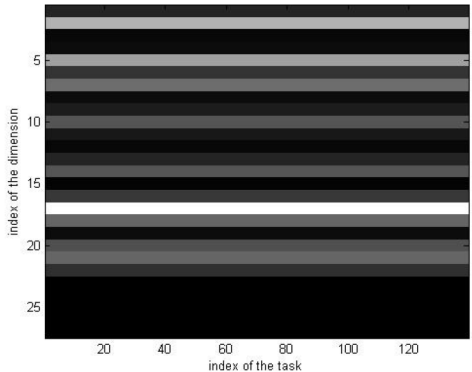}
	\end{center}
	\caption{Visualization of matrix $A_0$ learned on School dataset.}
\end{figure}

\begin{figure}[!tb]
	\label{fig_weightA}
	\begin{center}
		\includegraphics[width=0.4\textwidth]{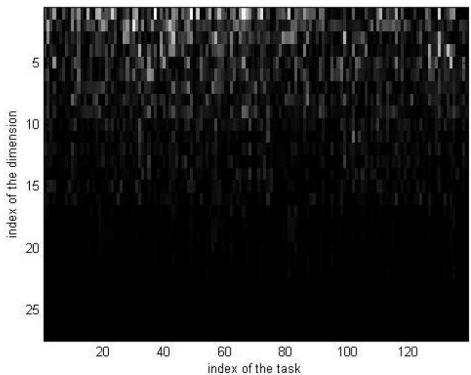}
	\end{center}
	\caption{Visualization of matrix $A$ learned on School dataset.}
\end{figure}

\section{Experiments} \label{sec:exp}
We show various experimental results and analyses on the experiments to demonstrate the effectiveness of our proposed multi-task learning method in this section. The comparison with several state-of-the-art multi-task learning algorithms further supports the  merits of our multi-task model and feature joint learning methods (MTMF). We compare our MTMF with two single-task learning methods - L2-norm regularized regression (L2-R) and L1-norm regularized regression (L1-R), as well as five state-of-the-art multi-task learning algorithms including trace norm regularized multi-task learning (TraceMT), low rank regularized multi-task learning with sparse structure (LowRankMT) \cite{Chen19}, convex multi-task feature learning(CMTL) \cite{argyriou4}, multi-task learning with a dirty model (MTDirty) \cite{Jalali12}, group sparse and low-rank regularized robust multi-task learning (SLMT) \cite{Chen14}. These five multi-task learning algorithms are representative methods of multi-task learning and the performance of them has been demonstrated to be promising on various datasets. The comparison with these methods can  sufficiently demonstrate the effectiveness of our proposed MTMF. The datasets used in our experiments are School dataset \footnote{http://ttic.uchicago.edu/~argyriou/code/.}, SARCOS dataset \footnote{http://www.gaussianprocess.org/gpml/data/.}, Isolet dataset \footnote{https://archive.ics.uci.edu/ml/datasets/ISOLET}, and MNIST dataset \footnote{http://yann.lecun.com/exdb/mnist/}.

\begin{table*}\tiny
\begin{center}
\caption{ Experimental results comparison on School dataset.}
\label{school dataset exp}
\begin{tabular}{|l|c|c|c|c|c|c|c|c|c|}
\hline
Measure & Training ratio & L2-R & L1-R & TraceMT & LowRankMT & CMTL & SLMT & MTDirty & MTMF\\
\hline
     & $10\%$ & $1.0398\pm0.0038$ & $1.0261\pm0.0132$ & $0.9359\pm0.0370$ & $0.9175\pm0.0261$ & $0.9413\pm0.0021$ & $0.9130\pm0.0039$ & $0.9543\pm0.0129$ & $\bm{0.7783\pm0.0082}$ \\
nMSE & $20\%$ & $0.8773\pm0.0043$ & $0.8754\pm0.0194$ & $0.8211\pm0.0032$ & $0.8126\pm0.0132$ & $0.8327\pm0.0039$ & $0.8055\pm0.0103$ & $0.8396\pm0.0142$ & $\bm{0.7432\pm0.0045}$\\
     & $30\%$ & $0.8171\pm0.0090$ & $0.8144\pm0.0091$ & $0.7870\pm0.0012$ & $0.7657\pm0.0091$ & $0.7922\pm0.0052$ & $0.7600\pm0.0032$ & $0.7985\pm0.0053$ & $\bm{0.7299\pm0.0064}$\\
\hline \hline
     & $10\%$ & $0.2713\pm+0.0023$ & $0.2682\pm0.0036$ & $0.2504\pm0.0102$ & $0.2419\pm0.0081$ & $0.2552\pm0.0032$ & $0.2330\pm0.0018$ & $0.2327\pm0.0031$ & $\bm{0.1898\pm0.0018}$ \\
aMSE & $20\%$ & $0.2303\pm0.0003$ & $0.2289\pm0.0051$ & $0.2156\pm0.0015$ & $0.2114\pm0.0041$ & $0.2131\pm0.0071$ & $0.2018\pm0.0025$ & $0.2048\pm0.0036$ & $\bm{0.1813\pm0.0010}$\\
     & $30\%$ & $0.2156\pm0.0021$ & $0.2137\pm0.0012$ & $0.2089\pm0.0012$ & $0.2011\pm0.0022$ & $0.1922\pm0.0102$ & $0.1822\pm0.0014$ & $0.1943\pm0.0016$ & $\bm{0.1776\pm0.0019}$\\
\hline
\end{tabular}
\end{center}
\end{table*}

\begin{table*}\tiny
\begin{center}
\caption{Experimental results comparison on SARCOS dataset.} \label{SARCOS dataset exp}
\begin{tabular}{|l|c|c|c|c|c|c|c|c|c|}
\hline
Measure & Training size &  L2-R & L1-R & TraceMT & LowRankMT & CMTL & SLMT & MTDirty & MTMF\\
\hline
     & $50$ & $0.2454\pm0.0260$ & $0.2337\pm0.0180$ & $0.2257\pm0.0065$ & $0.2127\pm0.0033$ & $0.2192\pm0.0016$ & $0.2123\pm0.0038$ & $0.1742\pm0.0178$ & $\bm{0.1640\pm0.0208}$ \\
nMSE & $100$ & $0.1821\pm0.0142$ & $0.1616\pm0.0027$ & $0.1531\pm0.0017$ & $0.1495\pm0.0023$ & $0.1568\pm0.0037$ & $0.1456\pm0.0138$ & $0.1274\pm0.0060$ & $\bm{0.1155\pm0.0215}$\\
     & $150$ & $0.1501\pm0.0054$ & $0.1469\pm0.0028$ & $0.1318\pm0.0053$ & $0.1236\pm0.0004$ & $0.1301\pm0.0034$ & $0.1245\pm0.0015$ & $0.1129\pm0.0039$ & $\bm{0.1057\pm0.0043}$\\
\hline \hline
     & $50$ & $0.1330\pm0.0143$ & $0.1228\pm0.0083$ & $0.1122\pm0.0064$ & $0.1073\pm0.0026$ & $0.1156\pm0.0011$ & $0.0982\pm0.0026$ & $0.0625\pm0.0063$ & $\bm{0.0588\pm0.0074}$ \\
aMSE & $100$ & $0.1053\pm0.0096$ & $0.0907\pm0.0023$ & $0.0805\pm0.0026$ & $0.0793\pm0.0047$ & $0.0852\pm0.0013$ & $0.0737\pm0.0083$ & $0.0458\pm0.0021$ & $\bm{0.0415\pm0.0023}$\\
     & $150$ & $0.0846\pm0.0045$ & $0.0822\pm0.0014$ & $0.0772\pm0.0023$ & $0.0661\pm0.0062$ & $0.0755\pm0.0025$ & $0.0674\pm0.0014$ & $0.0405\pm0.0011$ & $\bm{0.0379\pm0.0012}$\\
\hline
\end{tabular}
\end{center}
\end{table*}

\subsection{School dataset}
This dataset was collected to evaluate the effectiveness of schools by Inner London Education Authority.  It consists of 139 related tasks to predict the  examination scores of students from 139 secondary schools. The information of each student is encoded into a binary feature vector of 27 dimensions. There are totally 15362 samples. Single-task learning methods, such as L1-R and L2-R, learn these 139 tasks independently using their own data. All multi-task learning methods aim to improve the performance of these 139 tasks by uncovering the relatedness between tasks. The experimental settings follow previous works to fairly compare their performance. \par

Different ratios (10\%, 20\%, 30\% ) of training samples are randomly selected for training and the rest of samples are split into validation and test set. Consider the randomness of selection which may cause large variations in the results, we repeat all selections 10 times. All parameters are selected via the validation set. For all methods, the performance are evaluated by average mean squared error (aMSE) and normalized mean squared error (nMSE) which have been used in \cite{Gong13,Chen14}. The aMSE can be calculated through dividing the mean squared error by the variance of target vector and the nMSE can be calculated through dividing the mean squared error by the squared norm of target vector. \par

Table \ref{school dataset exp} gives the performance of all methods on School dataset. From the table, we can conclude that all multi-task learning methods can well uncover the relationships between tasks and improve the performance comparing to single-task learning methods. Another observation is that our proposed method performs the best with different training ratios. The improvement is especially obvious with a small amount of training samples, which indicates the success of our method to learn a new feature space and the strong ability of discovering latent relatedness between tasks. \par

To  analyse the properties of learned weight matrix $A_0 = [\underbrace{a_0,a_0,...,a_0}_T]$ and $A$,  we visualize them in  Fig. 2 and Fig. 3. The results are obtained using $20\%$ of the training samples. The zero values are denoted as black pixels in the figures. Most of the pixels in Figure 3 are black, which reveal the sparsity of the learned matrix $A$. A small subset of the features are shared across tasks corresponding to the 15 nonzero rows of matrix $A$. From Figure 2, we observe that $A_0$ is also a  sparse matrix. However, the features not used in matrix $A$ are appeared in matrix $A_0$, which means that our proposed MTMF can better utilize the information of the features. If we only use matrix A, all the tasks are forced to share some of the features without the utilization of other features. The relatedness between tasks becomes closer than they really are. $A_0$ helps all the tasks utilize more information from the features that are not shared through the matrix $A$. This is one of the reasons that our MTMF outperforms the CMTL.

\subsection{SARCOS dataset}

This dataset is used to learn the inverse dynamic of a SARCOS anthropomorphic robot arm. It aims to predict the seven joint torques using the provided 48933 samples described by a feature vector of 21 dimensions. In this experiment, we have seven tasks corresponding to predict these seven joint torques. Different amount of samples (50, 100, 150) are randomly selected as training data and 5000 samples are selected correspondingly as validation set and test set. The best parameters are selected on validation set for all methods. Consider the randomness of selection, we repeat all experiments 15 times and average performance is reported. \par

The comparison of experimental results between different methods is shown in Table \ref{SARCOS dataset exp}. Similar conclusions can be made to those of experiments on School dataset. Our proposed method can consistently outperform all other algorithms and all multi-task learning methods perform better than the two single-task learning methods. This further demonstrates the merits of multi-task learning and effectiveness of our method compared to other multi-task learning methods.

\subsection{Isolet dataset}
In this section, we conduct experiments on Isolet dataset from the UCI repository. It consists of 7797 pronunciation samples of the 26 English letters from 150 speakers. These speakers are split into five groups corresponding to five different tasks. The goal of each task is to predict the labels (1-26) of letters according to the pronunciation. In the experiment, labels of English letters are treated as regression values following the same setup as used in  \cite{Gong23}. Different ratio (15\%, 20\%, 25\%) of samples are randomly selected as training data and the rest is split into validation set and test set. All experiments are repeated 10 times and the best parameters are selected on validation set. We first reduce the dimensionality of the data to 100 using PCA. \par

The performance are reported in Table \ref{Isolet dataset exp}. Note that the two single-task learning methods L2-R and L1-R are not tested on Isolet dataset because of the bad performance on School and SARCOS datasets. Our proposed multi-task learning method outperforms other baselines obviously on this dataset, which proves the robustness of our method on various applications.

\begin{table*}
\begin{center}
\caption{Experimental results comparison on Isolet dataset.} \label{Isolet dataset exp}
\begin{tabular}{|l|c|c|c|c|c|c|c|}
\hline
Measure & Training ratio & TraceMT & LowRankMT & CMTL & SLMT & MTDirty & MTMF\\
\hline
     & $15\%$ & $0.6044\pm0.0154$ & $0.6307\pm0.0058$ & $0.7000\pm0.0106$ & $0.5987\pm0.0092$ & $0.6764\pm0.0112$ & $\bm{0.5691\pm0.0082}$ \\
nMSE & $20\%$ & $0.5705\pm0.0069$ & $0.6166\pm0.0093$ & $0.6491\pm0.0108$ & $0.5741\pm0.0078$ & $0.6344\pm0.0182$ & $\bm{0.5526\pm0.0046}$\\
     & $25\%$ & $0.5622\pm0.0086$ & $0.6011\pm0.0165$ & $0.6288\pm0.0049$ & $0.5635\pm0.0087$ & $0.6212\pm0.0299$ & $\bm{0.5498\pm0.0090}$\\
\hline \hline
     & $15\%$ & $0.1424\pm0.0035$ & $0.1486\pm0.0019$ & $0.1650\pm0.0029$ & $0.1411\pm0.0024$ & $0.1594\pm0.0029$ & $\bm{0.1314\pm0.0019}$ \\
aMSE & $20\%$ & $0.1343\pm0.0015$ & $0.1452\pm0.0022$ & $0.1528\pm0.0025$ & $0.1352\pm0.0017$ & $0.1494\pm0.0043$ & $\bm{0.1301\pm0.0012}$\\
     & $25\%$ & $0.1321\pm0.0025$ & $0.1412\pm0.0042$ & $0.1477\pm0.0017$ & $0.1324\pm0.0025$ & $0.1459\pm0.0076$ & $\bm{0.1292\pm0.0025}$\\
\hline
\end{tabular}
\end{center}
\end{table*}

\subsection{MNIST dataset}
We further study the effectiveness of our approach on a handwritten digit recognition dataset: MNIST. This dataset is composed of 60000 training examples and 10000 test examples. There are ten different handwritten digit numbers, corresponding to ten different binary classification tasks. Multi-way classification is treated as a multi-task learning problem, where each task is a classification task of one digit against all the other digits \cite{kang11,Amit07}.
We randomly select 500, 1000, and 1500 examples (50, 100, and 150 examples are selected from each digit number) from the 60000 training samples as training set and 1000 samples from the test samples to form the test sets. The dimensionality of images is reduced to 64 using PCA. All experiments are repeated 20 times and mean average precision (mAP) is reported.

\begin{table}
\begin{center}
\caption{Experimental Results comparison on MNIST.} \label{MNIST dataset exp}\tiny
\begin{tabular}{|l|c|c|c|}
\hline
Training size & 50 & 100 & 150\\
\hline
TraceMT & $0.8088\pm0.0118$ & $0.8297\pm0.0114$ & $0.8382\pm0.0111$ \\
\hline
LowRankMT & $0.7483\pm0.0260$ & $0.8088\pm0.0181$ & $0.8289\pm0.0192$ \\
\hline
CMTL & $0.8091\pm0.0108$ & $0.8343\pm0.0124$ & $0.8391\pm0.0115$ \\
\hline
SLMT & $0.7578\pm0.0165$ & $0.8144\pm0.0160$ & $0.8264\pm0.0147$ \\
\hline
MTDirty & $0.7955\pm0.0131$ & $0.8152\pm0.0128$ & $0.8202\pm0.0191$ \\
\hline
MTMF & $\bm{0.8180\pm0.0125}$ & $\bm{0.8394\pm0.0144}$ & $\bm{0.8484\pm0.0111}$ \\
\hline
\end{tabular}
\end{center}
\end{table}

The results on this dataset are shown in Table \ref{MNIST dataset exp}. We compare our MTMF method with five other multi-task regression learning methods. The results show that our proposed method outperforms the other five multi-task learning methods on the MNIST dataset.

\subsection{Analysis on p-values}
To further demonstrate that the proposed method is indeed statistically significantly better than the next best method, we present the p-values between our proposed method and the next best method in Table \ref{p-values exp}. The table includes six groups of experiments on the School dataset (nMSE: 10\%, 20\%, 30\%; aMSE: 10\%, 20\%, 30\%), SARCOS dataset (nMSE: 50, 100, 150; aMSE:50, 100, 150), Isolet dataset (nMSE: 15\%, 20\%, 25\%; aMSE: 15\%, 20\%, 25\%), and three groups of experiments on the MNIST dataset (AP: 50, 100, 150). We index the experiments for all the datasets from 1 to 6. From Table \ref{p-values exp}, our proposed method significantly outperforms the next best methods on the School dataset, Isolet dataset and MNIST dataset, as the p-values are substantially smaller than $0.05$. On the SARCOS dataset, our method does not perform significantly better than the next best method. However, the proposed method performs much better than all other methods. \par

The main reason that our proposed method outperforms other multi-task learning methods is that our proposed method considers the shared features and shared parameters simultaneously. Therefore, our proposed method can perform better if the data has both feature relatedness and model relatedness. Additionally, we can balance the importance between feature relatedness and model relatedness through tradeoff parameters $\gamma$ and $\beta$. Thus our model can degenerate to just share feature representations or share model. Consequently, our proposed model is more robust to various data.

\begin{table}
\tiny
\begin{center}
\caption{p-values between our proposed method and the next best method on all the datasets.}
\label{p-values exp}
\begin{tabular}{|l|c|c|c|c|}
\hline
Index Number & School dataset & SARCOS dataset & Isolet dataset & MNIST dataset\\
\hline
1 & $3.47\times 10^{-5}$ & $0.5963$ & $1.47\times 10^{-10}$ & $4.34 \times 10^{-5}$\\
\hline
2 & $2.86\times 10^{-8}$ & $0.4447$ & $2.87\times 10^{-6}$ & $2.59\times 10^{-2}$\\
\hline
3 & $2.64 \times 10^{-6}$ & $0.4245$ & $1.51 \times 10^{-6}$ & $1.33 \times 10^{-5}$\\
\hline
4 & $3.89 \times 10^{-6}$ & $0.5923$ & $2.42 \times 10^{-10}$ & -\\
\hline
5 & $2.26 \times 10^{-9}$ & $0.4436$ & $2.76 \times 10^{-6}$ & -\\
\hline
6 & $9.28 \times 10^{-5}$ & $0.4244$ & $1.62 \times 10^{-6}$ & - \\
\hline
\end{tabular}
\end{center}
\end{table}

\subsection{Sensitivity analysis on MTMF}
In this section, we conduct experiments to analyze the sensitivity of our proposed MTMF method. We will mainly discuss how the regularization parameters $\gamma$  and $\beta$ and the training size affect the performance of our MTMF method. All the experiments are conducted on the School dataset.

\begin{figure}[!tb]
\label{fig_sen size}
\begin{center}
\includegraphics[width=0.45\textwidth]{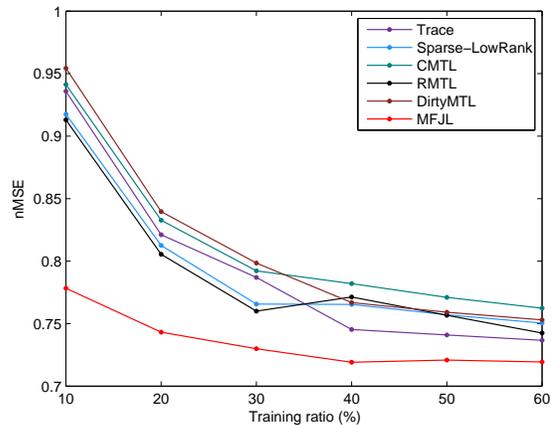}
\end{center}
\vspace{-3mm}
\caption{Sensitivity analysis on training size}
\end{figure}

\textbf{Analysis of the training ratio:} In these experiments, we randomly select $10\%, 20\%, 30\%, 40\%, 50\%$, and $60\%$ of the data as training sets and use the remaining data as test sets. We study how the training size affects the performance of MTMF. The experiments are repeated 10 times, and the regularization parameters $(\gamma,\beta)$ are selected through validation. The results are shown in Figure 4.
We can conclude that the proposed method outperforms the other methods significantly with consistent increase of training ratio.
It is also found that the performance of multi-task learning algorithms improves quicker when having a small amount of training samples and that the performance improves only slightly when the training ratio reaches a high level. It is consistent with the learning ability of multi-task learning. The relatedness between different tasks can provide more information to each task especially when the amount of training data is small. This results in a rapid increase in performance. However, the contribution of information from other tasks will decrease when task itself has sufficient training samples, which leads to a smaller increase in performance.

\textbf{Analysis of the regularization parameters:} We conduct experiments on the School dataset to analyze the sensitivity of the two regularization parameters. We randomly select $20\%$ of the data as training set and the remaining data as test set. For the sensitivity analysis of the parameter $\gamma$, we fix $\beta=1$ and vary the value of $\gamma$ as $\{1, 10, 100, 200, 500, 1000, 2000, 3000, 5000\}$. For the parameter $\beta$, we fix $\gamma=100$ and vary the value of $\beta$ as $\{10^{-5},10^{-4},10^{-3},10^{-2},10^{-1}, 1, 10, 50, 100\}$. The results are shown in Figure 5 and Figure 6. In Figure 5, we can see that the best performance by MFJL is obtained by setting $\gamma=1000$ when $\beta=1$ is fixed. From Figure 6, we see that the best performance by MTMF is obtained by setting the value of $\beta$ as a small value. Additionally, the performance of MFJL changes slightly when the value of $\beta$ is in the range of $[10^{-4}, 1]$. In general, MFJL performs well when the ratio $\frac{\gamma}{\beta}$ reaches a relatively high value (approximately 1000). This means that only a few features will be shared across tasks and that the central hyperplane $a_0$ will play an important role.

\begin{figure}[!tb]
\label{fig_sen gamma}
\begin{center}
\includegraphics[width=0.4\textwidth]{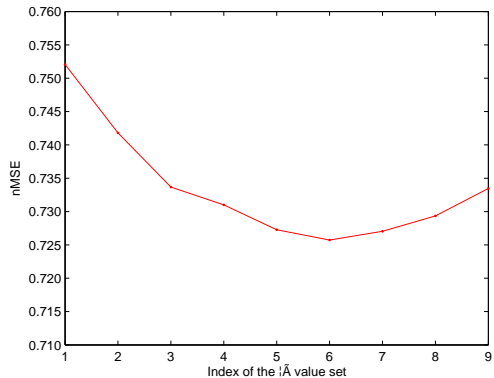}
\end{center}
\vspace{-3mm}
\caption{Sensitivity analysis on $\gamma$.}
\end{figure}

\begin{figure}[!tb]
\label{fig_sen beta}
\begin{center}
\includegraphics[width=0.4\textwidth]{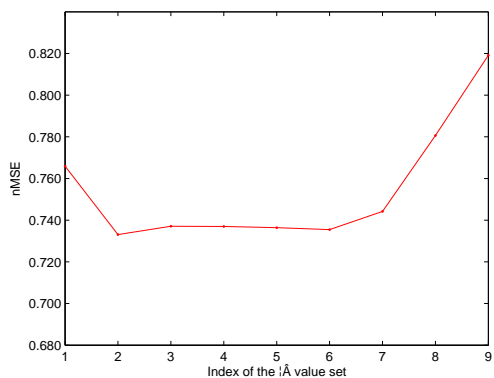}
\end{center}
\vspace{-3mm}
\caption{Sensitivity analysis on $\beta$.}
\end{figure}

\section{Conclusions and future work} \label{sec:conclusion}

In this paper, we summarize the defects of traditional multi-task learning methods and propose a novel multi-task learning framework, which learns shared latent feature representation and shared parameters jointly. The proposed method is introduced in detail and a new algorithm for optimizing the non-convex problem is proposed. Additionally, we theoretically demonstrate the merits of the proposed method compared to single-task learning and its strong ability to measure the relatedness between tasks. We conduct various experiments on four multi-task learning datasets and the results have demonstrated the effectiveness of the proposed method.  \par

In the future, we consider to extend the multi-task model and feature joint learning method into a more general framework. In this paper, the learned feature mapping matrix $U$ is an orthogonal matrix. It may be more efficient if the orthogonal matrix $U$ is replaced by a common matrix. Additionally, we make assumptions that all tasks share a common parameter, which is not suitable for some real-world cases. Considering this, we will attempt to automatically learn the relatedness between tasks and not make assumptions about the relatedness.

\appendices

\section{Proof of Theorem \ref{main}}

Before we provide the proof of Theorem \ref{main}, we need to introduce some used tools. We first give an introduction to the concentration inequality \cite{hoeffding1963probability}, which is better known as Hoeffding's inequality.
\begin{theorem}\label{hoeffding}
Let $x_1,\ldots,x_n$ be independent random variables with the range $[a_i,b_i]$ for $i=1,\ldots,n$. Let $S_n=\sum_{i=1}^{n}x_i$. Then, for any $\epsilon>0$, the following inequalities hold:
\[\text{Pr}\{S_n-ES_n\geq\epsilon\}\leq\exp\left(\frac{-2\epsilon^2}{\sum_{i=1}^{n}(b_i-a_i)^2}\right),\]
\[\text{Pr}\{ES_n-S_n\geq\epsilon\}\leq\exp\left(\frac{-2\epsilon^2}{\sum_{i=1}^{n}(b_i-a_i)^2}\right).\]
\end{theorem}
We then introduce the Rademacher complexity \cite{bartlett2003rademacher}, which is suitable to derive dimensionality-independent generalization bounds.
\begin{defn}
Let $X=\{x_1,\ldots,x_n\}\in \mathcal{X}^n$ be an independent distributed sample, and let $F$ be a function class on $\mathcal{X}$. Let $\sigma_1,\ldots,\sigma_n$ be independent Rademacher variables, which are uniformly distributed in $\{-1,1\}$. The empirical Rademacher complexity is defined as
\[\mathfrak{R}_n(F)=E_{\sigma}\sup_{f\in F}\frac{2}{n}\sum_{i=1}^{n}\sigma_if(x_i).\]
The Rademacher complexity is defined as
\[\mathfrak{R}(F)=E_x\mathfrak{R}_n(F).\]
\end{defn}

According to the symmetric distribution property of random variables, the following theorem \cite{mohri2012foundations} holds:
\begin{theorem}\label{symmRade}
Let
\[\Phi(X)=\sup_{f\in F}\frac{1}{n}\sum_{i=1}^{n}(E_xf(x)-f(x_i)).\]
Then,
\[E_x\Phi(X)\leq \mathfrak{R}(F).\]
\end{theorem}

Combining Theorem \ref{symmRade} and Hoeffding's inequality, we have the following:

\begin{theorem}[\cite{mohri2012foundations}]\label{rademachergener}
Let $F$ be an $[a,b]$-valued function class on $\mathcal{X}$, and $X=\{x_1,\ldots,x_n\}\in\mathcal{X}^n$. For any $\delta>0$, with a probability of at least $1-\delta$, we have
\begin{eqnarray*}
\sup_{f\in F}\left(E_xf(x)-\frac{1}{n}\sum_{i=1}^{n}f(x_i)\right)\leq \mathfrak{R}(F)+(b-a)\sqrt{\frac{\ln(1/\delta)}{2n}},
\end{eqnarray*}
or
\begin{eqnarray*}
\sup_{f\in F}\left(E_xf(x)-\frac{1}{n}\sum_{i=1}^{n}f(x_i)\right)\leq \mathfrak{R}_n(F)+3(b-a)\sqrt{\frac{\ln(2/\delta)}{2n}},
\end{eqnarray*}
\end{theorem}

The following property of Rademacher complexity \cite{bartlett2003rademacher} will help to construct the upper bound.
\begin{lema}\label{lemma}
If $\phi: \mathbb{R}\rightarrow\mathbb{R}$ is Lipschitz with constant $L$ and satisfies $\phi(0)=0$, then
\[\mathfrak{R}_n(\phi\circ F)\leq 2L\mathfrak{R}_n(F).\]
\end{lema}

\begin{lema}\label{lemma2}
Let
\[\mathfrak{R}_n(l\circ(A,a_0,U))=2E_\sigma\sup_{a_t,a_0,U}\sum_{t=1}^{T}\sum_{i=1}^{m_t}\sigma_{ti}l\left(y_{ti},\left<a_t+a_0,U^Tx_{ti}\right>\right),\]
where $\sigma_{ti}$ are Rademacher variables indexed by $t=1,\ldots,T$ and $i=1,\ldots,m_t$. We have
\[\mathfrak{R}_n(l\circ(A,a_0,U))\leq 2c\left(\sqrt{\frac{T}{\gamma}}+\sqrt{\frac{1}{\beta}}\right)\sqrt{\sum_{t=1}^{T}m_tS(X_t)},\]
where $S(X_t)=\text{tr}\left(\hat{\Sigma}(x_t)\right)=\frac{1}{m_t}\sum_{i=1}^{m_t}\|x_{ti}\|_2^2$ is the empirical covariance for the observations of the $t$-th task.
\end{lema}
\emph{Proof.} We have
\begin{eqnarray*}
&&\mathfrak{R}_n(l\circ(A,a_0,U))\\
&&=2E_\sigma\sup_{a_t,a_0,U}\sum_{t=1}^{T}\sum_{i=1}^{m_t}\sigma_{ti}l\left(y_{ti},\left<a_t+a_0,U^Tx_{ti}\right>\right)\\
&&=2E_\sigma\sup_{a_t,a_0,U}\sum_{t=1}^{T}\sum_{i=1}^{m_t}\sigma_{ti}l\left(y_{ti},\left<U(a_t+a_0),x_{ti}\right>\right)\\
&&\ \ \ (\text{Using Lemma \ref{lemma}})\\
&&\leq2cE_\sigma\sup_{a_t,a_0,U}\sum_{t=1}^{T}\sum_{i=1}^{m_t}\sigma_{ti}\left<U(a_t+a_0),x_{ti}\right>\\
&&\leq2cE_\sigma\sup_{a_t,U}\sum_{t=1}^{T}\sum_{i=1}^{m_t}\sigma_{ti}\left<Ua_t,x_{ti}\right>\\
&&\ \ \ +2cE_\sigma\sup_{a_0,U}\sum_{t=1}^{T}\sum_{i=1}^{m_t}\sigma_{ti}\left<Ua_0,x_{ti}\right>\\
&&=2cE_\sigma\sup_{a_t,U}\sum_{t=1}^{T}\left<Ua_t,\sum_{i=1}^{m_t}\sigma_{ti}x_{ti}\right>\\
&&\ \ \ +2cE_\sigma\sup_{a_0,U}\left<Ua_0,\sum_{t=1}^{T}\sum_{i=1}^{m_t}\sigma_{ti}x_{ti}\right>\\
&&\ \ \ (\text{Using H\"{o}lder's inequality})\\
&&\leq2cE_\sigma\sup_{a_t,U}\sqrt{\sum_{t=1}^{T}\left\|Ua_t\right\|_2^2}\sqrt{\sum_{t=1}^{T}\left\|\sum_{i=1}^{m_t}\sigma_{ti}x_{ti}\right\|_2^2}\\
&&\ \ \ +2cE_\sigma\sup_{a_0,U}\|Ua_0\|_2\left\|\sum_{t=1}^{T}\sum_{i=1}^{m_t}\sigma_{ti}x_{ti}\right\|_2\\
&&\ \ \ (\text{Since $U^TU=I$})\\
&&=2cE_\sigma\sup_{a_t}\sqrt{\sum_{t=1}^{T}\left\|a_t\right\|_2^2}\sqrt{\sum_{t=1}^{T}\left\|\sum_{i=1}^{m_t}\sigma_{ti}x_{ti}\right\|_2^2}\\
&&\ \ \ +2cE_\sigma\sup_{a_0}\|a_0\|_2\left\|\sum_{t=1}^{T}\sum_{i=1}^{m_t}\sigma_{ti}x_{ti}\right\|_2\\
\end{eqnarray*}

\begin{eqnarray*}
&&\leq2cE_\sigma\sup_{a_t}\|A\|_{2,1}\sqrt{\sum_{t=1}^{T}\left\|\sum_{i=1}^{m_t}\sigma_{ti}x_{ti}\right\|_2^2}\\
&&\ \ \ +2cE_\sigma\sup_{a_0}\|a_0\|_2\left\|\sum_{t=1}^{T}\sum_{i=1}^{m_t}\sigma_{ti}x_{ti}\right\|_2\\
&&\ \ \ \left(\text{Since $\|A\|_{2,1}^2\leq\frac{T}{\gamma}, \|a_0\|_2^2\leq\frac{1}{\beta}$}\right)\\
&&\leq \frac{2c\sqrt{T}}{\sqrt{\gamma}}E_\sigma\sqrt{\sum_{t=1}^{T}\left\|\sum_{i=1}^{m_t}\sigma_{ti}x_{ti}\right\|_2^2}+ \frac{2c}{\sqrt{\beta}}E_\sigma\left\|\sum_{t=1}^{T}\sum_{i=1}^{m_t}\sigma_{ti}x_{ti}\right\|_2\\
&&=\frac{2c\sqrt{T}}{\sqrt{\gamma}}E_\sigma\sqrt{\sum_{t=1}^{T}\left\|\sum_{i=1}^{m_t}\sigma_{ti}x_{ti}\right\|_2^2}+ \frac{2c}{\sqrt{\beta}}E_\sigma\sqrt{\left\|\sum_{t=1}^{T}\sum_{i=1}^{m_t}\sigma_{ti}x_{ti}\right\|_2^2}\\
&&\ \ \ (\text{Since the sqrt function is concave})\\
&&\leq
\frac{2c\sqrt{T}}{\sqrt{\gamma}}\sqrt{E_\sigma\sum_{t=1}^{T}\left\|\sum_{i=1}^{m_t}\sigma_{ti}x_{ti}\right\|_2^2}+\\
&&\frac{2c}{\sqrt{\beta}}\sqrt{E_\sigma\left\|\sum_{t=1}^{T}\sum_{i=1}^{m_t}\sigma_{ti}x_{ti}\right\|_2^2}\\
&&\ \ \ (\text{Since $\sigma_{ti}$ are independent and $E\sigma_{ti}=0,E\sigma_{ti}^2=1$})\\
&&\leq \frac{2c\sqrt{T}}{\sqrt{\gamma}}\sqrt{\sum_{t=1}^{T}\sum_{i=1}^{m_t}\left\|x_{ti}\right\|_2^2}+ \frac{2c}{\sqrt{\beta}}\sqrt{\sum_{t=1}^{T}\sum_{i=1}^{m_t}\left\|x_{ti}\right\|_2^2}\\
&&\leq 2c\left(\sqrt{\frac{T}{\gamma}}+\sqrt{\frac{1}{\beta}}\right)\sqrt{\sum_{t=1}^{T}m_tS(X_t)}.
\end{eqnarray*}\hfill$\blacksquare$

Theorem \ref{main} follows by combining Theorem \ref{rademachergener} and Lemma \ref{lemma2}.

\ifCLASSOPTIONcaptionsoff
  \newpage
\fi



%

\bibliographystyle{IEEEtran}
\bibliography{mybib}

%

%
%
%




\end{document}